\documentclass[11pt]{article}
\usepackage{color}
\usepackage[numbers]{natbib}
\usepackage{amstext, amssymb, amsthm, amsmath,bm,rotating,array,stmaryrd}
\usepackage{booktabs,url}
\usepackage[vlined,ruled,linesnumbered]{algorithm2e}
\usepackage[margin=1.2in]{geometry}
\usepackage{graphicx}
\usepackage{epstopdf}
\usepackage[lofdepth,lotdepth]{subfig}

\newcommand{\real}[1]{\mathrm{I \! R} \mathit{^{#1}}}
\newcommand{\trans}{^{\mbox{\tiny {\sf T}}}}

\newcommand{\sgn}{\mathrm{sgn}}

\newcommand{\argmin}{\mathrm{argmin}}

\newcommand{\Abf}{{\bm A}}

\newcommand{\Lbf}{{\bm L}}
\newcommand{\Mbf}{{\bm M}}

\newcommand{\Sbf}{{\bm S}}
\newcommand{\Ubf}{{\bm U}}
\newcommand{\Vbf}{{\bm V}}

\newcommand{\Xbf}{{\bm X}}
\newcommand{\Ybf}{{\bm Y}}

\newcommand{\abf}{{\bm a}}
\newcommand{\bbf}{{\bm b}}

\newcommand{\lbf}{{\bm l}}

\newcommand{\mbf}{{\bm m}}
\newcommand{\sbf}{{\bm s}}
\newcommand{\rbf}{{\bm r}}
\newcommand{\ubf}{{\bm u}}

\newcommand{\vbf}{{\bm v}}

\newcommand{\zerobf}{{\mathbf 0}}

\newcommand{\greekbold}[1]{\mbox{\boldmath $#1$}}

\newcommand{\Sigmabf}{\greekbold{\Sigma}}

\DeclareMathOperator{\Tr}{Tr}

\begin{document}
\title{Online Robust Principal Component Analysis with Change Point Detection}

\author{Wei Xiao$^\dagger$\thanks{Corresponding author.}, Xiaolin Huang$^\ddagger$, Jorge Silva$^\dagger$, Saba Emrani$^\dagger$ and Arin Chaudhuri$^\dagger$ \vspace{0.1in}\\
	\small{$^\dagger$\textit{SAS Institute, Cary, NC USA}}\\
	\small{$^\ddagger$\textit{Shanghai Jiao Tong University, Shanghai China}}\\
	\footnotesize{wei.xiao@sas.com \quad xiaolinhuang@sjtu.edu.cn \quad jorge.silva@sas.com \quad saba.emrani@sas.com \quad arin.chaudhuri@sas.com}}
\date{}

\baselineskip=20pt
\maketitle

\begin{abstract}
Robust PCA methods are typically batch algorithms which requires loading all observations into memory before processing. This makes them inefficient to process big data. In this paper, we develop an efficient online robust principal component methods, namely online moving window robust principal component analysis (OMWRPCA). Unlike existing algorithms, OMWRPCA  can successfully track not only slowly changing subspace but also abruptly changed subspace. By embedding hypothesis testing into the algorithm, OMWRPCA can detect change points of the underlying subspaces. Extensive simulation studies demonstrate the superior performance of OMWRPCA compared with other state-of-art approaches. We also apply the algorithm for real-time background subtraction of  surveillance video.
\end{abstract}

\section{Introduction}
Low-rank subspace models are powerful tools to analyze high-dimensional data from dynamical systems. Applications include tracking in radar and sonar \citep{krim1996two}, face recognition \citep{basri2003lambertian}, recommender system \citep{candes2009exact}, cloud removal in satellite images \citep{aravkin2014variational}, anomaly detection \citep{jiang2016online} and background subtraction for surveillance video \citep{candes2011robust,he2012incremental,bouwmans2014robust}. Principal component analysis (PCA) is the most widely used tool to obtain a low-rank subspace from high dimensional data. For a given rank $r$, PCA finds the $r$ dimensional linear subspace which minimizes the square error loss between the data vector and its projection onto that subspace. Although PCA is widely used, it has the drawback that it is not robust against outliers. Even with one grossly corrupted entry, the low-rank subspace estimated from classic PCA can be arbitrarily far from the true subspace. This shortcoming puts the application of the PCA technique into jeopardy for practical usage as outliers are often observed in the modern world's massive data. For example, data collected through sensors, cameras and websites are often very noisy and contains error entries or outliers.

Various versions of robust PCA have been developed in the past few decades, including \cite{de2003framework, candes2011robust, roweis1998algorithms, zhang2014novel, mccoy2011two}. Among them, the Robust PCA based on Principal Component Pursuit (RPCA-PCP) \citep{candes2011robust, wright2009robust} is the most promising one, as it has both good practical performance and strong theoretical performance guarantees. A comprehensive review of the application of Robust PCA based on Principal Component Pursuit in surveillance video can be found in \cite{bouwmans2014robust}. RPCA-PCP decomposes the observed matrix into a low-rank matrix and a sparse matrix by solving Principal Component Pursuit. Under mild condition, both the low-rank matrix and the sparse matrix can be recovered exactly.

Most robust PCA methods including RPCA-PCP are implemented in the batch manner. Batch algorithms not only require huge amount of memory storage as all observed data needs to be stored in the memory before processing, but also become too slow to process big data. An online version of robust PCA method is highly necessary to process incremental dataset, where the tracked subspace can be updated whenever a new sample is received. The online version of robust PCA can be applied to video analysis, change point detection, abnormal events detection and network monitoring. 

There are mainly three versions of online robust PCA proposed in the literature. They are Grassmannian Robust Adaptive Subspace Tracking Algorithm (GRASTA) \citep{he2011online}, Recursive Projected Compress Sensing (ReProCS) \citep{qiu2014recursive,guo2014online} and Online Robust PCA via Stochastic Optimization (RPCA-STOC) \citep{feng2013online}. GRASTA applies incremental gradient descent method on Grassmannian manifold to do robust PCA. It is built on Grassmannian Rank-One Update Subspace Estimation (GROUSE) \citep{balzano2010online}, a widely used online subspace tracking algirithm. GRASTA uses the $\ell_1$ cost function to replace the $\ell_2$ cost function in GROUSE. In each step, GRASTA updates the subspace with gradient descent. ReProCS is designed mainly to solve the foreground and background separation problem for video sequence. ReProCS recursively finds the low-rank matrix and the sparse matrix through four steps: 1. perpendicular projection of observed vector to low-rank subspace; 2. sparse recovery of sparse error through optimization; 3. recover low-rank observation; 4. update low-rank subspace. Both GRASTA and Prac-ReProCS can only handle slowly changing subspaces. RPCA-STOC is based on stochastic optimization for a close reformulation of the Robust PCA based on Principal Component Pursuit \citep{feng2013online}. In particular, RPCA-STOC splits the nuclear norm of the low-rank matrix as the sum of Frobenius norm of two matrices, and iteratively updates the low-rank subspace and finds the sparse vector. RPCA-STOC works only with stable subspace. 

Most of the  previously developed online robust PCA algorithms only deal with stable subspaces or slowly changing subspaces. However, in practice, the low-rank subspace may change suddenly, and the corresponding change points are usually of great interests. For example, in the application of background subtraction from video, each scene change corresponds to a change point in the underlying subspace. Another application is in failure detection of mechanical systems based on sensor readings from the equipment. When a failure happens, the underlying subspace is changed. Accurate identification of the  change point is useful for failure prediction. Some other applications include human activity recognition, intrusion detection in computer networks, etc. In this paper, we propose an efficient online robust principal component methods, namely online moving window robust principal component analysis (OMWRPCA), to handle both slowly changing and abruptly changed subspace. The method, which embeds hypothesis testing into online robust PCA framework, can accurately identify the change points of underlying subspace, and simultaneously estimate the low-rank subspace and sparse error. To the limit of the authors' knowledge, OMWRPCA is the first algorithm developed to be able to simultaneously detect change points and compute RPCA in an online fashion. It is also the first online RPCA algorithm that can handle both slowly changing and abruptly changed subspaces. 

The remainder of this paper is organized as follows. Section 2 gives the problem definition and introduces the related methods. The OMWRPCA algorithm is developed in Section 3. In Section 4, we compare OMWRPCA with RPCA-STOC via extensive numerial experiments. We also demonstrate the OMWRPCA algorithm with an application to a real-world video surveillance data. Section 5 concludes the paper and gives some discussion.

\section{Problem Definition and Related Works}
\subsection{Notation and Problem Definition}
We use bold letters to denote vectors and matrices. We use $\|\abf\|_1$ and $\|\abf\|_2$ to represent the $\ell_1$-norm and $\ell_2$-norm of vector $\abf$, respectively.  For an arbitrary real matrix $\Abf$, $\|\Abf\|_{F}$ denotes the Frobenius norm of matrix $\Abf$, $\|\Abf\|_{*}$ denotes the nuclear norm of matrix $\Abf$ (sum of all singular values) and $\|\Abf\|_1$ stands for the $\ell_1$-norm of matrix $\Abf$ where we treat $\Abf$ as a vector.

Let $T$ denote the number of observed samples, and $t$ is the index of the sample instance. We assume that our inputs are streaming observations $\mbf_t\in\real{m}$, $t=1,\ldots,T$, which can be decomposed as two parts $\mbf_t = \lbf_t + \sbf_t$. The first part $\lbf_t$ is a vector from a low-rank subspace $\Ubf_t$, and the second part $\sbf_t$ is a sparse error with support size $c_t$. Here $c_t$ represents the number of nonzero elements of $\sbf_t$, and $c_t=\sum_{i=1}^{m}I_{\sbf_t[i]\neq 0}$. The underlying subspace $U_t$ may or may not change with time $t$. When $\Ubf_t$ does not change over time, we say the data are generated from a stable subspace. Otherwise, the data are generated from a changing subspace. We assume $\sbf_t$ satisfy the sparsity assumption, i.e., $c_t\ll m$ for all $t=1,\ldots,T$. We use $\Mbf_t=[\mbf_1,\ldots,\mbf_t]\in\real{^{m\times t}}$ to denote the matrix of observed samples until time $t$. Let $\Mbf$, $\Lbf$, $\Sbf$ denotes $\Mbf_T$, $\Lbf_T$, $\Sbf_T$ respectively.

\subsection{Robust Principal Component Analysis}
Robust PCA based on Principal Component Pursuit (RPCA-PCP) \citep{candes2011robust, wright2009robust} is the most popular RPCA algorithm which decomposes the observed matrix $\Mbf$ into a low-rank matrix $\Lbf$ and a sparse matrix $\Sbf$ by solving Principal Component Pursuit:
\begin{equation}
\min_{\Lbf,\Sbf} \|\Lbf\|_{*}+\lambda\|\Sbf\|_1\text{ subject to }\Lbf+\Sbf=\Mbf.
\end{equation}
Theoretically, we can prove that $\Lbf$ and $\Sbf$ can be recovered exactly if (a) the rank of $\Lbf$ and the support size of $\Sbf$ are small enough; (b) $\Lbf$ is ``dense''; (c) any element in sparse matrix $\Sbf$ is nonzero with probability $\rho$ and 0 with probability $1-\rho$ independent over time \citep{candes2011robust,he2011online}. RPCA-PCP is often applied to estimate the low-rank matrix which approximates the observed data when the inputs are corrupted with gross but sparse errors.

Various algorithms have been developed to solve RPCA-PCP, including Accelerated Proximal Gradient (APG) \citep{lin2009fast} and Augmented Lagrangian Multiplier (ALM) \citep{lin2010augmented}. Bases on the experiment of \cite{candes2011robust}, ALM achieves higher accuracy than APG, in fewer iterations. We implemented ALM in this paper. Let $\mathcal{S}_{\tau}$ denote the shrinkage operator $S_{\tau}[x]=\sgn(x)\max(|x| - \tau, 0)$, and extend it to matrices by applying it to each element. Let $\mathcal{D}_{\tau}(\Xbf)$ denote the singular value thresholding operator given by $\mathcal{D}_{\tau}(\Xbf) = \Ubf\mathcal{S}_{\tau}(\Sigmabf)\Vbf^{*}$, where $\Xbf = \Ubf\Sigmabf\Vbf^*$ is any singular value decomposition. ALM is outlined in Algorithm \ref{algo:RPCA_PCP}. 

\begin{algorithm}[H]
	\DontPrintSemicolon
	Input: $\Mbf$ (observed data), $\lambda,\mu\in\real{}$ (regularization parameters)\;
	Initialize: $\Sbf_0=\Ybf_0 = \zerobf$.\;
	\While{not converged}{
		1) Compute $\Lbf^{(k+1)} \leftarrow \mathcal{D}_{\mu^{-1}}(\Mbf - \Sbf^{(k)} + \mu^{-1}\Ybf^{(k)})$;\;
		2) Compute $\Sbf^{(k+1)} \leftarrow \mathcal{S}_{\lambda\mu^{-1}}(\Mbf - \Lbf^{(k+1)} + \mu^{-1}\Ybf^{k})$;\;
		3) Compute $\Ybf^{(k+1)} \leftarrow \Ybf^{(k)} + \mu(\Mbf - \Lbf^{(k+1)} - \Sbf^{(k+1)});$\;
	}	
	\Return{$\Lbf$ (low-rank data matrix), $\Sbf$ (sparse noise matrix)}
	\caption{Principal Component Pursuit by Augumented Largrangian Multiplier \citep{lin2010augmented, candes2011robust}}
	\label{algo:RPCA_PCP}
\end{algorithm}

RPCA-PCP is a batch algorithm. It iteratively computes SVD with all data. Thus, it requires a huge amount of memory and does not scale well with big data.

\subsection{Moving Window Robust Principal Component Analysis}
In practice, the underlying subspace can change over time. This makes the RPCA-PCP infeasible to do after some time as the rank of matrix $\Lbf$ (includes all observed samples) will increase over time. One solution is Moving Window Robust Principal Component Analysis (MWRPCA) which applies a sliding window approach to the original data and iteratively computes batch RPCA-PCP using only the latest $n_{\mathrm{win}}$ data column vectors, where $n_{\mathrm{win}}$ is a user specified window size.

MWRPCA generally performs well for subspace tracking. It can do subspace tracking on both slowly changing subspace and abruptly changed subspace. However, MWRPCA often becomes too slow to deal with real-world big data problem. For example, suppose we want to track the subspace on an $\Mbf$ matrix with the size $200\times 10000$. If we chose a window size $n_{\mathrm{win}}=1000$, we need to run RPCA-PCP $9000$ times on matrices with size $200\times 1000$, which is computationally too expensive. 

\subsection{Robust PCA via Stochastic Optimization}
Feng et al. \citep{feng2013online} proposed an online algorithm to solve robust PCA based on Principal Component Pursuit. It starts by minimizing the following loss function 
\begin{equation}
\min_{\Lbf,\Sbf}\frac{1}{2}\|\Mbf-\Lbf-\Sbf\|_{F}^2+\lambda_1\|\Lbf\|_{*}+\lambda_2\|\Sbf\|_1,
\label{eq:stocRPCA}
\end{equation}
where $\lambda_1$, $\lambda_2$ are tuning parameters. The main difficulty in developing an online algorithm to solve the above equation is that the nuclear norm couples all the samples tightly. Feng et al. \citep{feng2013online} solve the problem by using an equivalent form of the nuclear norm following \cite{recht2010guaranteed}, which states that the nuclear norm for a matrix $\Lbf$ whose rank is upper bounded by $r$ has an equivalent form
\begin{equation}
\|\Lbf\|_{*} = \inf_{\Ubf\in\real{m\times r},\Vbf\in\real{r\times n}}
\left\{\frac{1}{2}\|\Ubf\|^2_{F}+\frac{1}{2}\|\Vbf\|^2_{F}:\,\Lbf=\Ubf\Vbf
\right\}.
\label{eq:splitL}
\end{equation}
Substituting $\Lbf$ by $\Ubf\Vbf$ and plugging \eqref{eq:splitL} into \eqref{eq:stocRPCA}, we have 
\begin{equation}
\min_{\Ubf\in\real{m\times r},\Vbf\in\real{n\times r},\Sbf}
\frac{1}{2}\|\Mbf-\Ubf\Vbf-\Sbf\|_{F}^2+\frac{\lambda_1}{2}(\|\Ubf\|^{2}_{F}+ \|\Vbf\|^2_{F})+\lambda_2\|\Sbf\|_1,
\label{eq:stocRPCA2}
\end{equation}
where $\Ubf$ can be seen as the basis for low-rank subspace and $\Vbf$ represents the coefficients of observations with respect to the basis. They then propose their RPCA-STOC algorithm which minimizes the empirical version of loss function \eqref{eq:stocRPCA2} and processes one sample per time instance \citep{feng2013online}. Given a finite set of samples $\Mbf_t=[\mbf_1,\ldots,\mbf_t]\in\real{m\times t}$, the empirical version of loss function \eqref{eq:stocRPCA2} at time point $t$ is
\begin{equation*}
f_t(\Ubf)=\frac{1}{t}\sum_{i=1}^{t}\ell(\mbf_i,\Ubf)+\frac{\lambda_1}{2t}\|\Ubf\|^{2}_{F},
\label{eq:empirical_loss_stocRPCA}
\end{equation*}
where the loss function for each sample is defined as
\begin{equation*}
\ell(\mbf_i,\Ubf)\triangleq\min_{\vbf,\sbf}\frac{1}{2}\|\mbf_i-\Ubf\vbf-\sbf\|_{2}^{2}+\frac{\lambda_1}{2}\|\vbf\|_{2}^{2}+\lambda_2\|\sbf\|_1.
\end{equation*}
Fixing $\Ubf$ as $\Ubf_{t-1}$, $\vbf_t$ and $\sbf_t$ can be obtained by solving the optimization problem
\begin{equation*}
(\vbf_t,\sbf_t)=\underset{\vbf,\sbf}\argmin\frac{1}{2}\|\mbf_t-\Ubf\vbf-\sbf\|_{2}^{2}+\frac{\lambda_1}{2}\|\vbf\|_{2}^{2}+\lambda_2\|\sbf\|_1.
\end{equation*}
Assuming $\{\vbf_i,\sbf_i\}_{i=1}^{t}$ are known, the basis $\Ubf_t$ can be updated by minimizing the following function
\begin{equation*}
g_t(\Ubf)\triangleq\frac{1}{t}\sum_{i=1}^{t}\left(\frac{1}{2}\|\mbf_i-\Ubf\vbf_i\|_{2}^{2}+\frac{\lambda_1}{2}\|\vbf_i\|_{2}^{2}+\lambda_2\|\sbf_i\|_1\right)+\frac{\lambda_1}{2t}\|\Ubf\|_{F}^2,
\label{eq:loss_stoc_rpca}
\end{equation*}
which gives the explicit solution
\begin{equation*}
\Ubf_t = \left[\sum_{i=1}^{t}(\mbf_i-\sbf_i)\vbf_i\trans\right]\left[\left(\sum_{i=1}^t\vbf_i\vbf_i\trans\right)+\lambda_1I\right]^{-1}.
\end{equation*}
In practice, $\Ubf_t$ can be quickly updated by block-coordinate descent with warn restarts \citep{feng2013online}.
The RPCA-STOC algorithm is summarized in Algorithm~\ref{algo:ORPCA}.\\ 
\begin{algorithm}[H]
	\DontPrintSemicolon
	Input: $\{\mbf_1,\ldots,\mbf_T\}$(observed data which are revealed sequentially), $\lambda_1,\lambda_2\in\real{}$(tuning parameters), $T$(number of iterations);\;
	Initialize: $\Ubf_0 = \zerobf^{m\times r}$, $A_0 = \zerobf^{r\times r}$, $B_0 = \zerobf^{m\times r}$;\;
	\For{t=1 to T}{
		1) Reveal the sample $\mbf_t$.\;
		2) Project the new sample:
		\begin{equation*}
		(\vbf_t,\sbf_t)\leftarrow \argmin\frac{1}{2}\|\mbf_t-\Ubf_{t-1}\vbf-\sbf\|_{2}^{2}+\frac{\lambda_1}{2}\|\vbf\|_{2}^{2}+\lambda_2\|\sbf\|_1.
		\end{equation*}\;
		3) $A_t\leftarrow A_{t-1}+\vbf_t\vbf_t\trans$, $B_t\leftarrow B_{t-1}+(\mbf_t-\sbf_t)\vbf_t\trans$.\;
		4) Compute $\Ubf_t$ with $\Ubf_{t-1}$ as warm restart using Algorithm \ref{algo:basisupdate2}:
		\begin{equation*}
		\Ubf_t\triangleq\argmin\frac{1}{2}\Tr[\Ubf\trans(A_t+\lambda_1I)\Ubf]-\Tr(\Ubf\trans B_t).
		\end{equation*}
	}	
	\Return{$\Lbf_T=\{\Ubf_1\vbf_1,\ldots,\Ubf_T\vbf_T\}$ (low-rank data matrix), $\Sbf_T=\{\sbf_1,\ldots,\sbf_T\}$ (sparse noise matrix)}
	\caption{RPCA-STOC algorithm \citep{feng2013online}}
	\label{algo:ORPCA}
\end{algorithm}

\begin{algorithm}[H]
	\caption{Fast Basis Update\citep{feng2013online}}
	\label{algo:basisupdate2}
	\DontPrintSemicolon
	Input: $\Ubf=[\ubf_1,\ldots,\ubf_r]\in\real{m\times r},\,A=[\abf_1,\ldots,\abf_r]\in\real{r\times r}$, and $B=[\bbf_1,\ldots,\bbf_r]\in\real{r\times r}$; $\tilde{A}\leftarrow A +\lambda_1I$.\;
	\For{j=1 to r} {
		\begin{align*}
		\tilde{\ubf}_j \leftarrow& \frac{1}{\tilde{A}[j,j]}(\bbf_j-\Ubf\tilde{\abf}_j)+\ubf_j,\\
		\ubf_j \leftarrow& \frac{1}{\max(\|\tilde{\ubf}_j\|_2,1)}\tilde{\ubf}_j.
		\end{align*}		
	}
	\Return{$\Ubf$}
\end{algorithm}

\section{Online Moving Window RPCA}

\subsection{Basic Algorithm}
One limitation of RPCA-STOC is that the method assumes a stable subspace, which is generally a too restrict assumption for applications such as failure detection in mechanical systems, intrusion detection in computer networks, fraud detection in financial transaction and background subtraction for surveillance video. At time $t$, RPCA-STOC updates the basis of subspace by minimizing an empirical loss which involves all previously observed samples with equal weights. Thus, the subspace we obtain at time $t$ can be viewed as an ``average'' subspace of observations from time 1 to $t$. This is clearly not desirable if the underlying subspace is changing over time. We propose a new method which combines the idea of moving window RPCA and RPCA-STOC to track changing subspace. At time $t$, the proposed method updates $\Ubf_t$ by minimizing an empirical loss based only on the most recent $n_{\mathrm{win}}$ samples. Here $n_{\mathrm{win}}$ is a user specified window size and the new empirical loss is defined as 
\begin{equation*}
g^*_t(\Ubf)\triangleq\frac{1}{n_{\mathrm{win}}}\sum_{i=t-n_{\mathrm{win}}+1}^{t}\left(\frac{1}{2}\|\mbf_i-\Ubf\vbf_i\|_{2}^{2}+\frac{\lambda_1}{2}\|\vbf_i\|_{2}^{2}+\lambda_2\|\sbf_i\|_1\right)+\frac{\lambda_1}{2n_{\mathrm{win}}}\|\Ubf\|_{F}^2.
\label{eq:loss_omwrpca}
\end{equation*}
We call our new method Online Moving Window RPCA (OMWRPCA). The biggest advantage of OMWRPCA comparing with RPCA-STOC is that OMWRPCA can quickly update the subspace when the underlying subspace is changing. It has two other minor differences in the details of implementation compared with RPCA-STOC. In RPCA-STOC, we assume the dimension of subspace ($r$) is known. In OMWRPCA, we estimate it by computing batch RPCA-PCP with burn-in samples. The size of the burn-in samples $n_{\mathrm{burnin}}$ is another user specified parameter. We also use the estimated $\Ubf$ from burn-in samples as our initial guess of $\Ubf$ in OMWRPCA, while $\Ubf$ is initialized as zero matrix in RPCA-STOC. Specifically, in the initialization step, batch RPCA-PCP is computed on burn-in samples to get rank $r$, basis of subspace $\Ubf_0$, $A_0$ and $B_0$. For simplicity, we also require $n_\mathrm{{win}}\leq n_\mathrm{{burnin}}$. Details are given as follows.
\begin{enumerate}
	\item Compute RPCA-PCP on burn-in samples $\Mbf^{b}$, and we have $\Mbf^{b} = \Lbf^{b}+\Sbf^{b}$, where $\Lbf^{b}=[\lbf_{-n_\mathrm{burnin}+1},\ldots,\lbf_{0}]$ and $\Sbf^{b}=[\sbf_{-n_\mathrm{burnin}+1},\ldots,\sbf_{0}]$. 
	\item From SVD, we have $\Lbf^{b}=\hat{\Ubf}\hat{\Sigma}\hat{\Vbf}$, where $\hat{\Ubf}\in\real{m\times r}$, $\hat{\Sigmabf}\in\real{r\times r}$, and $\hat{\Vbf}\in\real{r\times n_{\mathrm{burnin}}}$.  
	\item $\Ubf_0 = \hat{\Ubf}\hat{\Sigmabf}^{1/2}\in\real{m\times r}$, $A_0 = \sum_{i=-(n_\mathrm{{win}}-1)}^{0}\vbf_i\vbf_i\trans\in\real{r\times r}$ and $B_0 = \sum_{i=-(n_\mathrm{{win}}-1)}^{0}(\mbf_i - \sbf_{i})\vbf_i\trans\in\real{m\times r}$;
\end{enumerate}  
OMWRPCA is summarized in Algorithm~\ref{algo:OMWRPCA}.

\begin{algorithm}[H]
	\DontPrintSemicolon
	Input: $\{\mbf_1,\ldots,\mbf_T\}$(observed data which are revealed sequentially), $\lambda_1,\lambda_2\in\real{}$(regularization parameters), $T$(number of iterations); $\Mbf^{b}=[\mbf_{-(n_\mathrm{{burnin}}-1)},\ldots,\mbf_{0}]$ (burn-in samples);\;
	Initialize: Compute batch RPCA-PCP on burn-in samples $\Mbf^{b}$ to get $r$, $\Ubf_0$, $A_0$ and $B_0$.\;  
	\For{t=1 to T}{
		1) Reveal the sample $\mbf_t$.\;
		2) Project the new sample:
		\begin{equation*}
		(\vbf_t,\sbf_t)\leftarrow\argmin\frac{1}{2}\|\mbf_t-\Ubf_{t-1}\vbf-\sbf\|_{2}^{2}+\frac{\lambda_1}{2}\|\vbf\|_{2}^{2}+\lambda_2\|\sbf\|_1.
		\end{equation*}\;
		3) $A_t\leftarrow A_{t-1}+\vbf_t\vbf_t\trans-\vbf_{t-n_{\mathrm{win}}}\vbf_{t-n_{\mathrm{win}}}\trans$, $B_t\leftarrow B_{t-1}+(\mbf_t-\sbf_t)\vbf_t\trans-(\mbf_{t-n_{\mathrm{win}}}-\sbf_{t-n_{\mathrm{win}}})\vbf_{t-n_{\mathrm{win}}}\trans$.\;
		4) Compute $\Ubf_t$ with $\Ubf_{t-1}$ as warm restart using Algorithm \ref{algo:basisupdate2}:
		\begin{equation*}
		\Ubf_t\triangleq\argmin\frac{1}{2}\Tr[\Ubf\trans(A_t+\lambda_1I)\Ubf]-\Tr(\Ubf\trans B_t).
		\end{equation*}
	}	
	\Return{$\Lbf_T=\{\Ubf_1\vbf_1,\ldots,\Ubf_T\vbf_T\}$ (low-rank data matrix), $\Sbf_T=\{\sbf_1,\ldots,\sbf_T\}$ (sparse noise matrix)}
	\caption{Online Moving Window RPCA}
	\label{algo:OMWRPCA}
\end{algorithm}
  
\subsection{Change Point Detection in Online Moving Window RPCA with Hypothesis Testing}
One limitation of the basic OMWRPCA algorithm is that it can only deal with slowly changing subspace. When the subspace changes suddenly, the basic OMWRPCA algorithm will fail to update the subspace correctly. This is because when the new subspace is dramatically different from the original subspace, the subspace may not be able to be updated in an online fashion or it may take a while to finish the updates. One example is that in the case the new subspace is in higher dimension than the original subspace, basic OMWRPCA algorithm can never be able to update the subspace correctly as the rank of the subspace is fixed to a constant. Similar drawback is shared by the majority of the other online RPCA algorithms which have been previously developed. Furthermore, an online RPCA algorithm which can identify the change point is very desirable since the change point detection is very crucial for subsequent identification, moreover, we sometimes are more interested to pinpoint the change points than to estimate the underlying subspace. Thus, we propose another variant of our OMWRPCA algorithm to simultaneously detect change points and compute RPCA in an online fashion. The algorithm is robust to dramatic subspace change. We achieve this goal by embedding hypothesis testing into the original OMWRPCA algorithm. We call the new algorithm OMWRPCA-CP. 

\begin{figure}[ht]
	\centering
	\includegraphics[width=1\textwidth]{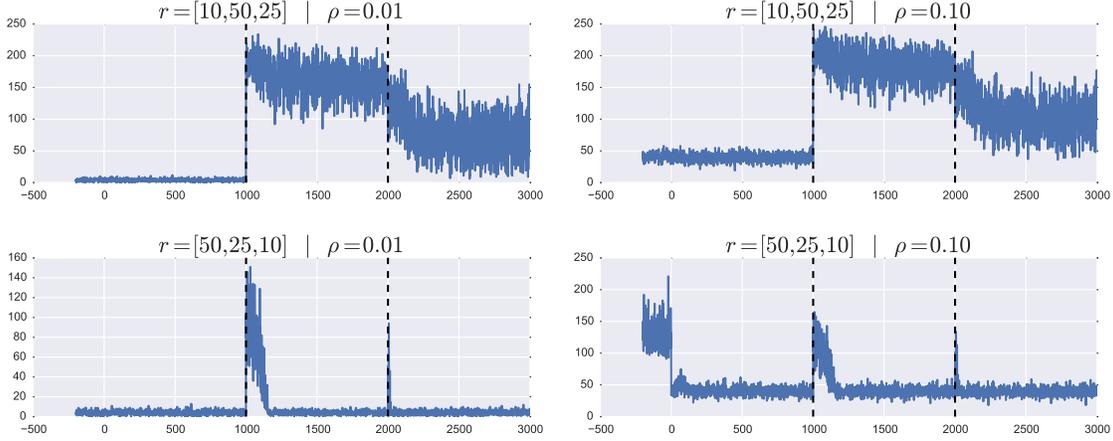}
	\caption{Line plots of support sizes of estimated sparse vector from OMWRPCA based on four simulated data.}
	\label{plot:ct_plot_sim3}
\end{figure}

OMWRPCA-CP is based on a very simple yet important observation. That is when the new observation can not be modeled well with the current subspace, we will often result in an estimated sparse vector $\hat{\sbf}_t$ with an abnormally large support size from the OMWRPCA algorithm. So by monitoring the support size of the sparse vector computed from the OMWRPCA algorithm, we can identify the change point of subspace which is usually the start point of level increases in the time series of $\hat{c}_t$. Figure~\ref{plot:ct_plot_sim3} demonstrates this observation on four simulated cases. In each plot, support sizes of estimated sparse vectors are plotted. For each case, burn-in samples are between the index -200 and 0, and two change points are at $t=1000$ and 2000. The time line is separated to three pieces with change points. In each piece, we have a constant underlying subspace, and the rank is given in $r$. The elements of the true sparse vector have a  probability of $\rho$ to be nonzero, and are independently generated. The dimension of the samples is 400. We choose window size 200 in OMWRPCA. Theoretically, when the underlying subspace is accurately estimated, the support sizes of the estimated sparse vectors $\hat{c}_t$ should be around 4 and 40 for $\rho=0.01$ case and $\rho=0.1$ case, respectively. For the upper two cases, $\hat{c}_t$ blows up after the first change point and never return to the normal range. This is because after we fit RPCA-PCP on burn-in samples, the dimension of $\Ubf_t$ is fixed to 10. Thus, the estimated subspace $\Ubf_t$ can never approximate well the true subspace in later two pieces of the time line as the true subspaces have larger dimensions. On the other hand, for the lower two cases, $\hat{c}_t$ blows up immediately after each change point, and then drops back to the normal range after a while when the estimated subspace has converged to the true new subspace.

We give an informal proof of why the above phenomenon happens. Assume we are at time point $t$ and the observation is $\mbf_t=\Ubf_t\vbf_t+\sbf_t$. We estimate ($\hat{\vbf}_t$,\,$\hat{\sbf}_t$) by computing 		
\begin{equation*}
(\hat{\vbf}_t,\hat{\sbf}_t)=\argmin\frac{1}{2}\|\mbf_t-\hat{\Ubf}_{t-1}\vbf-\sbf\|_{2}^{2}+\frac{\lambda_1}{2}\|\vbf\|_{2}^{2}+\lambda_2\|\sbf\|_1,
\end{equation*}
where $\hat{\Ubf}_{t-1}$ is the estimated subspace at time point $t-1$. The above optimization problem does not have an explicit solution and can be iteratively solved by fixing $\sbf_t$, solving $\vbf_t$ and fixing $\vbf_t$, solving $\sbf_t$. We here approximate the solution with one step update from a good initial point $\hat{\sbf}_t=\sbf_t$. We have $\hat{\vbf}_t=(\hat{\Ubf}_{t-1}\trans\hat{\Ubf}_{t-1}+\lambda_1I)^{-1}\hat{\Ubf}_{t-1}\trans\Ubf_t\vbf_t$, $\hat{\sbf}_t=\mathcal{S}_{\lambda_2}\left\{\sbf_t + \left[I-\hat{\Ubf}_{t-1}(\hat{\Ubf}_{t-1}\trans\hat{\Ubf}_{t-1} + \lambda_1 I)^{-1}\hat{\Ubf}_{t-1}\right]\Ubf_{t}\vbf_t\right\}$. When $\lambda_1$ is small, we approximately have $\hat{\vbf}_t=(\hat{\Ubf}_{t-1}\trans\hat{\Ubf}_{t-1})^{-1}\hat{\Ubf}_{t-1}\trans\Ubf_t\vbf_t$ and $\hat{\sbf}_t=\mathcal{S}_{\lambda_2}\{\sbf_t + \mathcal{P}_{\hat{\Ubf}_{t-1}^{\bot}}\Ubf_{t}\vbf_t\}$, where $\mathcal{P}_{\hat{\Ubf}_{t-1}^{\bot}}$ represents the projection to the orthogonal complement subspace of $\hat{\Ubf}_{t-1}$. A good choice of $\lambda_1$ is in the order of $O\left(1/\sqrt{\max(m,n_{\mathrm{win}})}\right)$, and $\lambda_1$ is small when $\max(m,n_{\mathrm{win}})$ is large. We will discuss the tuning of parameters in more details later. Under the mild assumption that the nonzero elements of $\sbf_t$ are not too small $(\min_{i\in\{i:\,\sbf_t[i]\neq 0\}}\sbf_t[i] > \lambda_2)$, we have $\hat{c}_t = c_t$ when $\Ubf_t=\hat{\Ubf}_{t-1}$, and $\hat{c}_t = c_t + \#\{i:\, \sbf_t[i]= 0\text{ and }\mathcal{P}_{\hat{\Ubf}_{t-1}^{\bot}}\Ubf_{t}\vbf_t[i]>\lambda_2\} > c_t$ when $\max_{i\in\{i:\,\sbf_t[i]= 0\}}\left(\mathcal{P}_{\hat{\Ubf}_{t-1}^{\bot}}\Ubf_{t}\vbf_t[i]\right) > \lambda_2$. This analysis also provides us a clear mathematical description of the term ``abruptly changed subspace'', i.e., for a piecewise constant subspace, we say that a change point exists at time $t$  when the vector $\mathcal{P}_{\Ubf_{t-1}^{\bot}}\Ubf_t\vbf_t$  is not close to zero.

We develop a change point detection algorithm by monitoring $\hat{c}_t$. The algorithm determines that a change point exists when it finds $\hat{c}_t$ is abnormally high for a while. Specifically, the users can provide two parameters $N_{\mathrm{check}}$ and $\alpha_{\mathrm{prop}}$. When in $N_{\mathrm{check}}$ consecutive observations, the algorithm finds more than $\alpha_{\mathrm{prop}}$ observations with abnormally high $\hat{c}_t$, the algorithm decides that a change point exists in these $N_{\mathrm{check}}$ observations and traces back to find the change point. We refer the cases when the underlying estimated subspace approximates the true subspace well as ``normal'', and ``abnormal'' otherwise.  All the collected information of $\{\hat{c}_j\}_{j=1}^{t-1}$ from the normal period is stored in $H_c\in\real{m+1}$, where the $(i+1)$-th element of $H_c$ denotes the number of times that $\{\hat{c}_j\}$ equals $i$. When we compute $\hat{c}_t$ from a new observation $\mbf_t$, based on $H_c$ we can flag this observation as a normal one or an abnormal one via hypothesis testing. We compute p-value $p=\sum_{i=\hat{c}_t+1}^{m+1}H_c[i]/\sum_{i=0}^{m+1}H_c[i]$, which is the probability that we observe a sparse vector with at least as many nonzero elements as the current observation under the hypothesis that this observation is normal. If $p\leq\alpha$, we flag this observation as abnormal. Otherwise, we flag this observation as normal. Here $\alpha$ is a user specified threshold. We find $\alpha=0.01$ works very well. Algorithm 5 displays OMWRPCA-CP in pseudocode. The key steps are
\begin{enumerate}
	\item Initialize $H_c$ as a zero vector $\zerobf^{m+1}$. Initialize the buffer list for flag $B_f$ and the buffer list for count $B_c$ as empty lists. In practice, $B_f$ and $B_c$ can be implemented as queue structures (first in first out). 
	\item Collect $n_{\mathrm{burnin}}$ samples to get $\Mbf^{b}$. Run RPCA-PCP on $\Mbf^{b}$ and start OMWRPCA.
	\item For the first $n_{\mathrm{cp-burnin}}$ observations of OMWRPCA, we wait for the subspace of OMWRPCA algorithm to become stable. $H_c$, $B_f$ and $B_c$ remain unchanged.
	\item For the next $n_{\mathrm{test}}$ observations of OMWRPCA, we calculate $\hat{c}_t$ from $\hat{\sbf}_t$, and update $H_c$ ($H_c[\hat{c}_t + 1]\leftarrow H_c[\hat{c}_t + 1] + 1$). $B_f$ and $B_c$ remain unchanged. Hypothesis testing is not done in this stage, as the sample size of the hypothesis test is small ($\sum_{i=0}^{m+1}H_c[i]\leq n_{\mathrm{test}}$) and the test will not be accurate.
	\item Now we can do hypothesis testing on $\hat{c}_t$ with $H_c$. We get the flag $f_t$ for the $t$th observation, where $f_t=1$ if the $t$th observation is abnormal, and $f_t=0$ if the $t$th observation is normal. Append $f_t$ and $\hat{c}_t$ to $B_f$ and $B_c$, respectively. Denote the size of buffer $B_f$ or equivalently the size of buffer $B_c$ as $n_b$. If $n_b$ satisfies $n_b=N_{\mathrm{check}}+1$, pop one observation from both $B_f$ and $B_c$ on the left-hand side. Assuming the number popped from $B_c$ is $c$, we then update $H_c$ with $c$ ($H_c[c+1]\leftarrow H_c[c+1] + 1$). If $n_b < N_{\mathrm{check}}+1$, we do nothing. Thus, the buffer size $n_b$ is always kept within $N_{\mathrm{check}}$, and when the buffer size reaches $N_{\mathrm{check}}$, it will remain in $N_{\mathrm{check}}$. Buffers $B_f$ contains the flag information of most recent $N_{\mathrm{check}}$ observations and can be used to detect change points.    
	\item If $n_b$ equals $N_{\mathrm{check}}$, We compute $n_{\mathrm{abnormal}}=\sum_{i=1}^{N_{\mathrm{check}}}{B_f[i]}$ and compare it with $\alpha_{\mathrm{prop}}N_{\mathrm{check}}$. If $n_{\mathrm{abnormal}} \geq \alpha_{\mathrm{prop}}N_{\mathrm{check}}$, we know a change point exists in the most recent $N_{\mathrm{check}}$ observations. We then do a simple loop over the list $B_f$ to find the change point, which is detected as the first instance of $n_\mathrm{positive}$ consecutive abnormal cases. For example, suppose we find that $B_f[i]$ which corresponding to time points $t_0$, $t_0+1$, $\ldots$, $t_0+n_\mathrm{positive}-1$ are the first instance of $n_\mathrm{positive}$ consecutive abnormal cases. The change point is then determined as $t_0$. Here $n_\mathrm{positive}$ is another parameter specified by the user. In practice, we find $n_\mathrm{positive}=3$ works well. After the change point is identified, OMWRPCA-CP can restart from the change point.           
\end{enumerate}

\begin{algorithm}[ht]
	\DontPrintSemicolon
	Input: $\{\mbf_1,\ldots,\mbf_T\}$(observed data which are revealed sequentially), $\lambda_1,\lambda_2\in\real{}$(regularization parameters)\;
	$t=1$; $c_p$ is initialized as an empty list. \; 
	$t^*\leftarrow \min(t+n_\mathrm{{burnin}}-1,\,T)$; Compute batch RPCA-PCP on burn-in samples $\Mbf^{b}=[\mbf_{t},\ldots,\mbf_{t^*}]$ to get $r$, $\Ubf_{t^*}$, $A_{t^*}$ and $B_{t^*}$; $t\leftarrow t^*+1$.\;
	$t_{\mathrm{start}}\leftarrow t$; $H_c\leftarrow \zerobf^{m+1}$; $B_f$ and $B_c$ are initialized as empty lists. \; 
	\While{$t\leq T$}{
		1) Reveal the sample $\mbf_t$\;
		2) Project the new sample:
		\begin{equation*}
		(\vbf_t,\sbf_t)\leftarrow\argmin\frac{1}{2}\|\mbf_t-\Ubf_{t-1}\vbf-\sbf\|_{2}^{2}+\frac{\lambda_1}{2}\|\vbf\|_{2}^{2}+\lambda_2\|\sbf\|_1.
		\end{equation*}\;
		3) $A_t\leftarrow A_{t-1}+\vbf_t\vbf_t\trans-\vbf_{t-n_{\mathrm{win}}}\vbf_{t-n_{\mathrm{win}}}$, $B_t\leftarrow B_{t-1}+(\mbf_t-\sbf_t)\vbf_t\trans-(\mbf_{t-n_{\mathrm{win}}}-\sbf_{t-n_{\mathrm{win}}})\vbf_{t-n_{\mathrm{win}}}\trans$.\;
		4) Compute $\Ubf_t$ with $\Ubf_{t-1}$ as warm restart using Algorithm \ref{algo:basisupdate2}:
		\begin{equation*}
		\Ubf_t\triangleq\argmin\frac{1}{2}\Tr[\Ubf\trans(A_t+\lambda_1I)\Ubf]-\Tr(\Ubf\trans B_t).
		\end{equation*}\;
		5) Compute $c_t \leftarrow \sum_{i=1}^{m}\sbf_t[i]$.\;
		6) \uIf{$t < t_{\mathrm{start}}+n_{\mathrm{cp-burnin}}$}{
			Go to next loop;\;
		}
		\uElseIf{$t_{\mathrm{start}}+n_{\mathrm{cp-burnin}}\leq t < t_{\mathrm{start}}+n_{\mathrm{cp-burnin}}+n_{\mathrm{test}}$}{
			$H_c[c_t + 1]\leftarrow H_c[c_t + 1] + 1$; Go to next loop;\;		   	
		}
		\Else{
			Do hypothesis testing on $c_t$ with $H_c$, and get p-value $p$; $f_t\leftarrow I_{p\leq \alpha}$.\;
			Append $c_t$ and $f_t$ to $B_c$ and $B_f$, respectively.\;
			\If{size of $B_f = N_{\mathrm{check}} + 1$}{
				Pop one element from both $B_c$ and $B_f$ at left-hand side ($c\leftarrow \mathrm{Pop(B_c)}$, $f\leftarrow \mathrm{Pop(B_f)}$); Update $H_c$ ($H_c[c + 1]\leftarrow H_c[c + 1] + 1$).\;  
			}
			\If{size of $B_f = N_{\mathrm{check}}$}{
				Compute $n_{\mathrm{abnormal}} \leftarrow \sum_{i=1}^{N_{\mathrm{check}}}{B_f[i]}$\;
				\If{$n_{\mathrm{abnormal}} \geq \alpha_{\mathrm{prop}}N_{\mathrm{check}}$}{
					Find change point $t_0$ by looping over $B_f$; Append $t_0$ to $c_p$.\; 
					$t\leftarrow t_0$; Jump to step 3.\;					
				}
				\Else{Go to next loop;\;}	
			}	
		}
	}	
		\Return{$c_p$ (list of all change points)}
		\caption{Online Moving Window RPCA with Change Point Detection}
		\label{algo:OMWRPCA-CP}
\end{algorithm}

Good choice of tuning parameters is the key to the success of the proposed algorithms. ($\lambda_1$, $\lambda_2$) can be chosen based on cross-validation on the order of $O\left(1/\sqrt{\max(m,n_{\mathrm{win}})}\right)$. A rule of thumb choice is $\lambda_1=1/\sqrt{\max(m,n_{\mathrm{win}})}$ and $\lambda_2=100/\sqrt{\max(m,n_{\mathrm{win}})}$. $N_{\mathrm{check}}$ needs to be kept smaller than $n_{\mathrm{win}}/2$ to avoid missing a change point, and not too small to avoid generating false alarms. $\alpha_{\mathrm{prop}}$ can be chosen based on user's prior-knowledge. Sometimes the assumption that the support size of sparse vector $\sbf_t$ remains stable and much smaller than $m$ is violated in real world data. For example, in video surveillance data, the foreground may contain significant variations over time. In this case, we can add one additional positive tuning parameter $n_{\mathrm{tol}}$ and change the formula of p-value to $p=\sum_{i=\hat{c}_t-n_{\mathrm{tol}}+1}^{m+1}H_t[i]/\sum_{i=0}^{m+1}H_t[i]$. This change can make the hypothesis test more conservative, and force the algorithm not detecting too many change points (false alarms). 

It is easy to prove that OMWRPCA and OMWRPCA-CP (ignore the burn-in samples training) have the same computational complexity as STOC-RPCA. The computational cost of each new observation is $O(mr^2)$,  which is independent of the sample size and linear in the dimension of observation \citep{feng2013online}. In contrast, RPCA-PCP computes an SVD and a thresholding operation in each iteration with the computational complexity $O(nm^2)$. Based on the experiment of \cite{feng2013online}, the proposed method are also more efficient than other online RPCA algorithm such as GRASTA. The memory costs of OMWRPCA and OMWRPCA-CP are $O(mr)$ and $O(mr+N_{\mathrm{check}})$ respectively. In contrast, the memory cost of RPCA-PCP is $O(mn)$, where $n\gg r$.  Thus, the proposed methods are well suitable to process big data.

Last, current version of OMWRPCA-CP does hypothesis tests based on all historical information of $\hat{c}_t$. We can easily change $H_c$ to a queue structure and store only recent history of $\hat{c}_t$, where the user can specify how long of the history they would like to trace back. This change makes the algorithm detect a change point only by comparing with recent history. We do not pursue this variation in this paper.

\section{Numerical Experiments and Applications}

In this section, we compare OMWRPCA with RPCA-STOC via extensive numerial experiments and an application to a real-world video surveillance data. To make a fair comparison between OMWRPCA and RPCA-STOC, we estimate the rank $r$ in RPCA-STOC  with burn-in samples following the same steps as OMWRPCA. We implement all algorithms in python and the code is available at Github address \url{https://github.com/wxiao0421/onlineRPCA.git}.

\subsection{Simulation Study 1: stable subspace}
The simulation study has a setting similar to \cite{feng2013online}. The observations are generated through $\Mbf=\Lbf+\Sbf$, where $\Sbf$ is a sparse matrix with a  fraction of $\rho$ non-zero elements. The non-zero elements of $\Sbf$ are randomly chosen and generated from a uniform distribution over the interval of $[-1000,\,1000]$. The low-rank subspace $\Lbf$ is generated as a product  $\Lbf=\Ubf\Vbf$, where the sizes of $\Ubf$ and $\Vbf$ are $m\times r$ and $r \times T$ respectively. The elements of both $\Ubf$ and $\Vbf$ are i.i.d. samples from the $\mathcal{N}(0,1)$ distribution. Here $\Ubf$ is the basis of the constant subspace with dimension $r$.  We fix $T=5000$ and $m=400$. A burn-in samples $\Mbf^{b}$ with the size $400\times 200$ is also generated. We have four settings $(r,\,\rho)=(10, 0.01)$, $(10,0.1)$, $(50,0.01)$, and $(50,0.1)$. For each setting, we run 50 replications. We choose the following parameters in all simulation studies, $\lambda_1=1/\sqrt{400}$, $\lambda_2=100/\sqrt{400}$, $n_{\mathrm{win}}=200$, $n_{\mathrm{burnin}}=200$, $n_{\mathrm{cp-burnin}}=200$, $n_{\mathrm{test}}=100$, $N_{\mathrm{check}}=20$, $\alpha_{\mathrm{prop}}=0.5$, $\alpha=0.01$, $n_{\mathrm{positive}}=3$ and $n_{\mathrm{tol}}=0$.

We compare three different methods, STOC-RPCA, OMWRPCA and OMWRPCA-CP. Three criteria are applied:
\begin{align*}
\mathrm{ERR}_{\mathrm{L}} =& \|\hat{\Lbf}-\Lbf\|/\|\Lbf\|_F;\\
\mathrm{ERR}_{\mathrm{S}} =& \|\hat{\Sbf}-\Sbf\|/\|\Sbf\|_F;\\
\mathrm{F}_{\mathrm{S}} =& \#\{(\hat{\Sbf}\neq\zerobf)\neq(\Sbf\neq\zerobf)\}/(mT).
\end{align*}
Here $\mathrm{ERR}_{\mathrm{L}}$ is the relative error of the low-rank matrix $\Lbf$, $\mathrm{ERR}_{\mathrm{S}}$ is the relative error of the sparse matrix $\Sbf$, and $\mathrm{F}_{\mathrm{S}}$ is the  proportion of correctly identified elements in $\Sbf$. 

\begin{figure}[ht]
	\centering
	\includegraphics[width=1\textwidth]{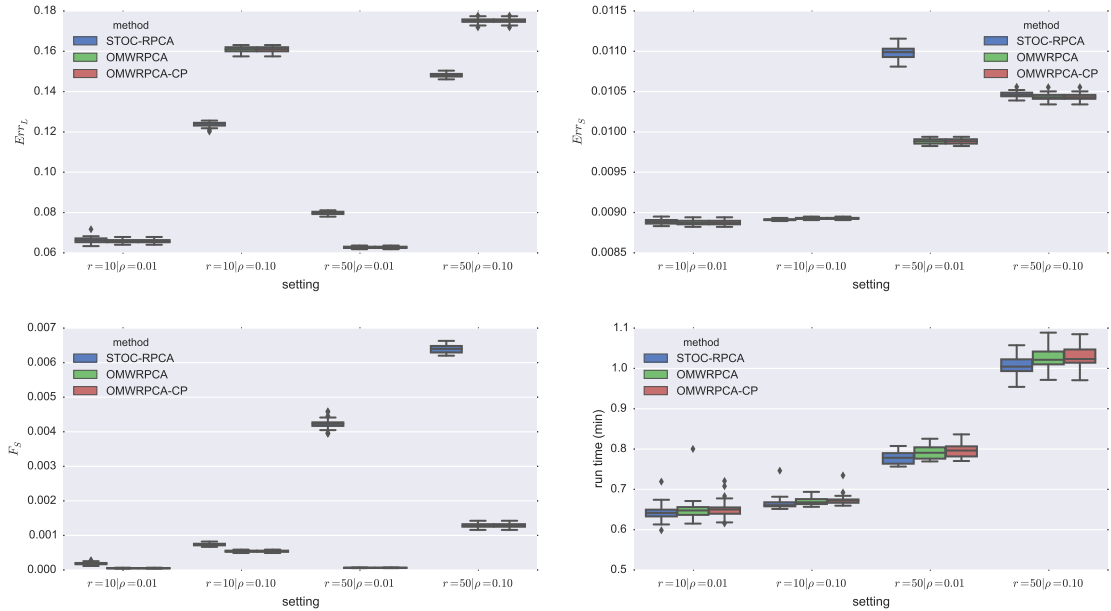}
	\caption{Box plots of $\mathrm{ERR}_{\mathrm{L}}$, $\mathrm{ERR}_{\mathrm{S}}$, $\mathrm{F}_{\mathrm{S}}$ and running times under the simulation study of stable subspace. For each setting, the box plots of STOC-RPCA, OMWRPCA and OMWRPCA-CP are arranged from left to right.}
	\label{plot:boxplot_sim1}
\end{figure}

Box plots of $\mathrm{ERR}_{\mathrm{L}}$, $\mathrm{ERR}_{\mathrm{S}}$, $\mathrm{F}_{\mathrm{S}}$ and running times are shown in Figure~\ref{plot:boxplot_sim1}. OMWRPCA-CP has the same result as OMWRPCA, and no change point is detected with OMWRPCA-CP in all replications. All three methods STOC-RPCA, OMWRPCA and OMWRPCA-CP has comparable performance on $\mathrm{ERR}_{\mathrm{S}}$ and running times. STOC-RPCA has slightly better performance on $\mathrm{ERR}_{\mathrm{L}}$ when $\rho=0.1$. OMWRPCA and OMWRPCA-CP has slightly better performance on $\mathrm{F}_{\mathrm{S}}$. All methods take approximately the same amount of time to run, and are all very fast (less than 1.1 minutes per replication for all settings). In contrast, MWRPCA takes around 100 minutes for the settings $(r,\,\rho)=(10, 0.01)$, $(10,0.1)$, $(50,0.01)$, and more than 1000 minutes for the setting $(r,\,\rho)=(50, 0.1)$.

\subsection{Simulation Study 2: slowly changing subspace}
We adopt almost all setting of Simulation Study 1 except that we make the underlying subspace $\Ubf$ change linearly over time. We first generate $\Ubf_0\in\real{m\times r}$ with i.i.d. samples from the $\mathcal{N}(0,1)$ distribution. We generate burnin samples $\Mbf^b$ based on $\Ubf_0$. We make $\Ubf$ slowly change over time by adding new matrics $\{\tilde{\Ubf}_k\}_{k=1}^{K}$ generated independently with i.i.d. $\mathcal{N}(0,1)$ elements to the first $r_0$ columns of $\Ubf$, where $\tilde{\Ubf}_k\in\real{m\times r_0}$, $k=1,\ldots,K$ and $K=T/T_p$. We choose $T_p=250$. Specifically, for $t=T_p*i + j$, where $i=0,\ldots,K-1$, $j=0,\ldots,T_p-1$, we have 
\begin{equation*}
\Ubf_t[:,1:r_0] = \Ubf_0[:,1:r_0]  + \sum_{1\leq k \leq i}\tilde{\Ubf}_k + \frac{j}{n_p}\tilde{\Ubf}_{i+1},\quad \Ubf_t[:,(r_0+1):r] = \Ubf_0[:,(r_0+1):r]. 
\end{equation*}
We set $r_0=5$ in this simulation.

\begin{figure}[ht]
	\centering
	\includegraphics[width=1\textwidth]{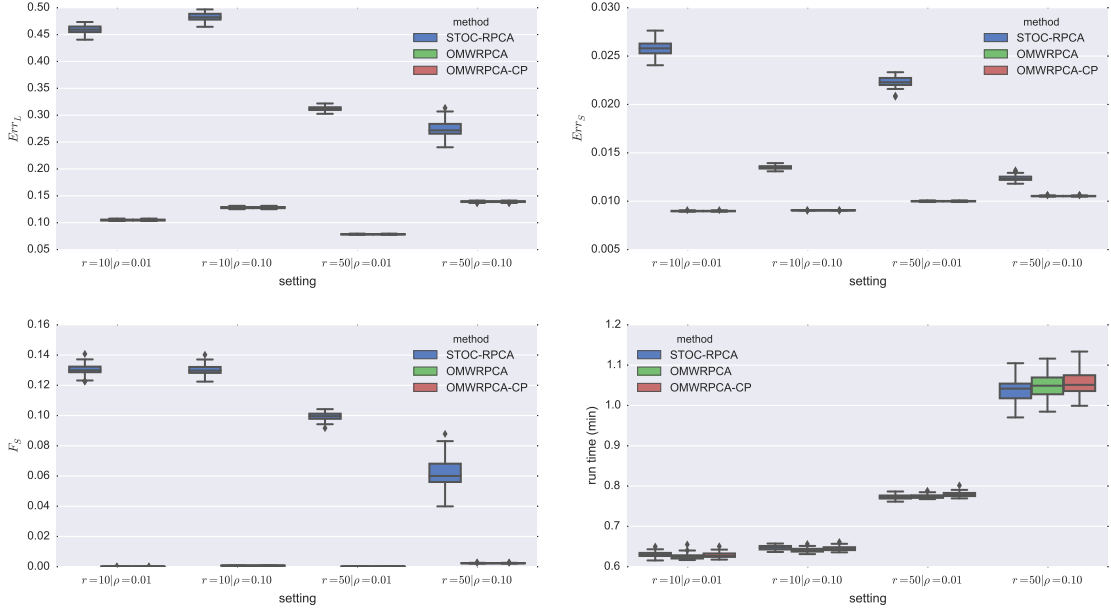}
	\caption{Box plots of $\mathrm{ERR}_{\mathrm{L}}$, $\mathrm{ERR}_{\mathrm{S}}$, $\mathrm{F}_{\mathrm{S}}$ and running times under the simulation study of slowly changing subspace. For each setting, the box plots of STOC-RPCA, OMWRPCA and OMWRPCA-CP are arranged from left to right.}
	\label{plot:boxplot_sim2}
\end{figure}

\begin{figure}[ht]
	\centering
	\includegraphics[width=1\textwidth]{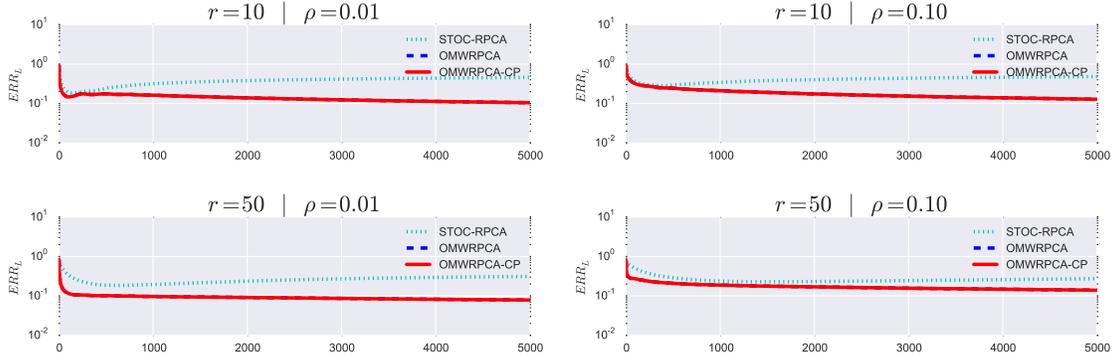}
	\caption{Line plots of average $\mathrm{ERR}_{\mathrm{L}}$ over time $t$ under the simulation study of slowly changing subspace.}
		\label{plot:errorL_sim2}
\end{figure}

Box plots of $\mathrm{ERR}_{\mathrm{L}}$, $\mathrm{ERR}_{\mathrm{S}}$, $\mathrm{F}_{\mathrm{S}}$ and running times are shown in Figure~\ref{plot:boxplot_sim2}. OMWRPCA-CP has the same result as OMWRPCA, and no change point is detected with OMWRPCA-CP in all replications. OMWRPCA and OMWRPCA-CP have better performance on all three criteria $\mathrm{ERR}_{\mathrm{L}}$, $\mathrm{ERR}_{\mathrm{S}}$, $\mathrm{F}_{\mathrm{S}}$ comparing with STOC-RPCA as we pointed out before that STOC-RPCA is not able to efficiently track changing subspace. In Figure~\ref{plot:errorL_sim2}, we plot the average of $\mathrm{ERR}_{\mathrm{L}}$ across all replications as a function of number of observations to investigate the progress of performance across different methods. It shows that the performance of STOC-RPCA deteriorates over time, while OMWRPCA and OMWRPCA-CP's performance is quite stable. 

\subsection{Simulation Study 3: slowly changing subspace with change points}
We adopt almost all setting of Simulation Study 2 except that we add two change points at time point 1000 and 2000, where the underlying subspace $\Ubf$ is changed and generated completely independently. These two change points cut the time line into three pieces, and the start of the subspace $\Ubf$ for each piece has rank $\rbf=(r_1,r_2,r_3)\trans$, where $r_i$ is the rank of $\Ubf$ for $i$th piece. We consider three settings of $\rbf$, $(10,10,10)\trans$, $(50,50,50)\trans$ and $(10,50,25)\trans$. We let the subspace $\Ubf$ slowly changing over time in each piece as we assumed in Simulation Study 2, where $T_p=250$. We set $T=3000$. 

\begin{figure}[ht]
	\centering
	\includegraphics[width=1\textwidth]{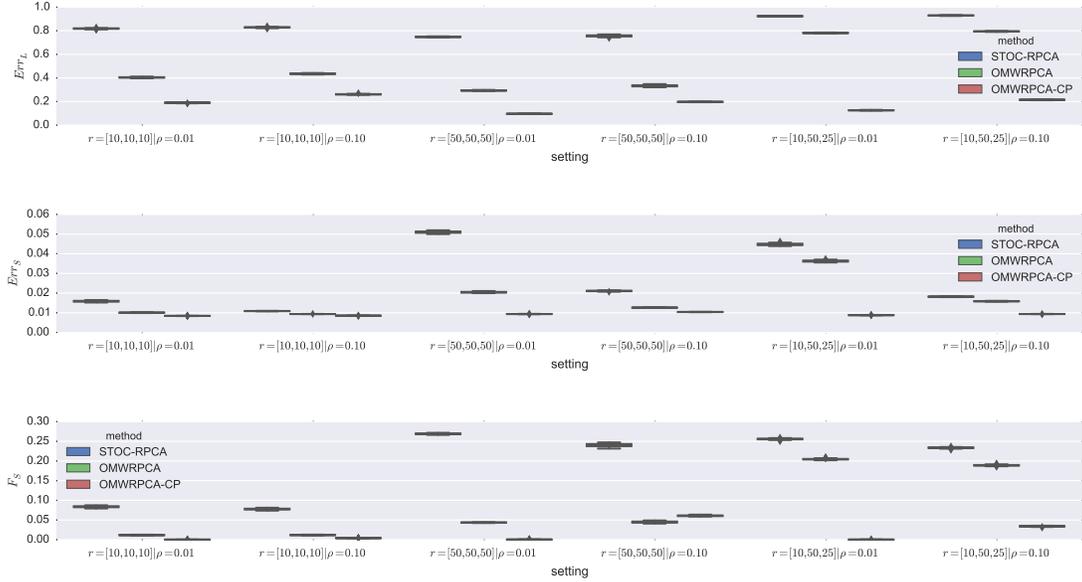}
	\caption{Box plots of $\mathrm{ERR}_{\mathrm{L}}$, $\mathrm{ERR}_{\mathrm{S}}$ and $\mathrm{F}_{\mathrm{S}}$ under the simulation study of slowly changing subspace with change points. For each setting, the box plots of STOC-RPCA, OMWRPCA and OMWRPCA-CP are arranged from left to right.}
	\label{plot:boxplot_sim4}
\end{figure}

\begin{figure}[ht]
	\centering
	\includegraphics[width=1\textwidth]{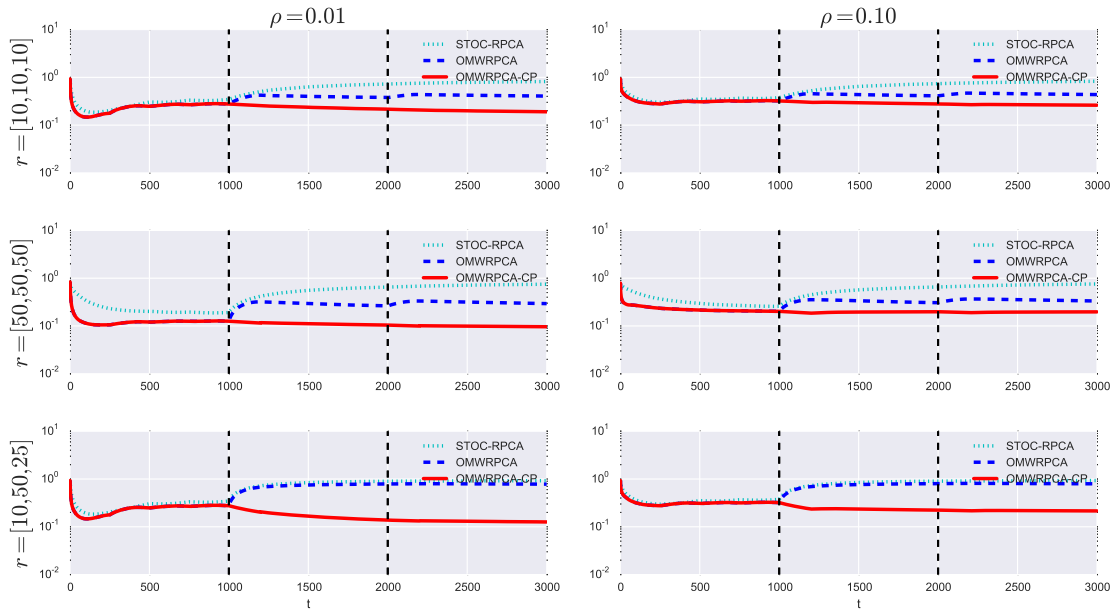}
	\caption{Line plots of average $\mathrm{ERR}_{\mathrm{L}}$ over time $t$ under the simulation study of suddenly changed subspace. Two change points are at time point 1000 and 2000 are marked by vertical lines.}
	\label{plot:errorL_sim4}
\end{figure}

\begin{figure}[ht]
	\centering
	\includegraphics[width=1\textwidth]{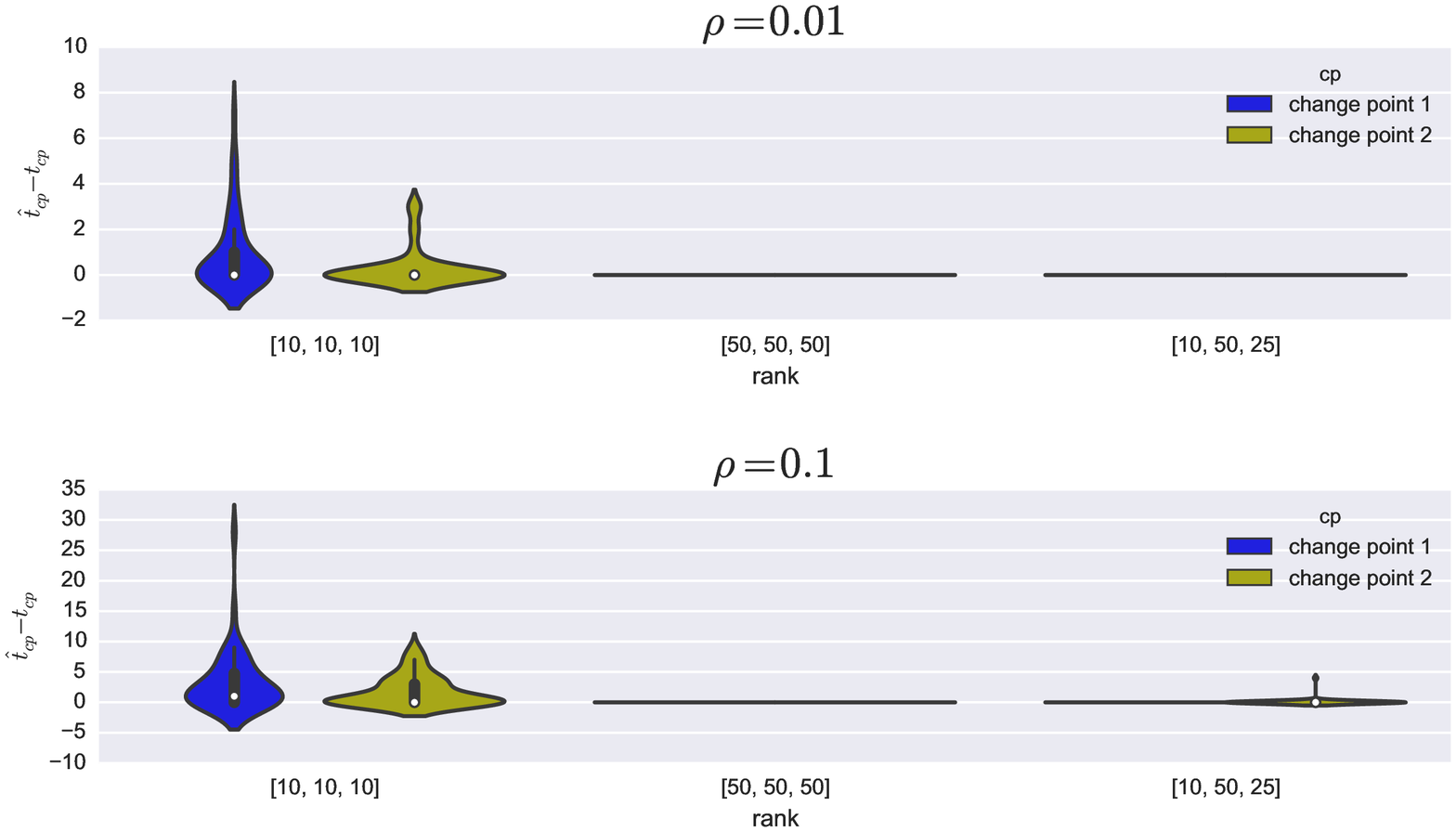}
	\caption{Violin plots of the difference between detected change points and true change points under the simulation study of suddenly changed subspace. }
	\label{plot:deviance_cp_sim4}
\end{figure}

Box plots of $\mathrm{ERR}_{\mathrm{L}}$, $\mathrm{ERR}_{\mathrm{S}}$ and $\mathrm{F}_{\mathrm{S}}$ at time point $T$ are shown in Figure~\ref{plot:boxplot_sim4}. Under almost all settings, OMWRPCA outperforms STOC-RPCA, and OMWRPCA-CP has the best performance of all three methods. In Figure~\ref{plot:errorL_sim4}, we plot the average of $\mathrm{ERR}_{\mathrm{L}}$ across all replications as a function of time $t$ to investigate the progress of performance across different methods. The performance of both STOC-RPCA and OMWRPCA deteriorate quickly after the first change point $t=1000$, while the performance of OMWRPCA-CP is stable over time. This indicates that only OMWRPCA-CP can track subspace correctly under the scenario of suddenly changed subspace. Furthermore we find OMWRPCA-CP correctly identify two change points for all replications over all settings. The distribution of the difference between the detected change points and the true change points ($\delta_{cp}=\hat{t}_{cp} - t_{cp})$ is shown in Figure~\ref{plot:deviance_cp_sim4}. We find $\delta_{cp}$ is almost always 0 when the subspaces before and after the change point are highly distinguishable, which represents the cases $\rbf=(50,50,50)\trans$ and $(10,50,25)\trans$. For the cases when the change of subspace is not dramatic ($\rbf=(10,10,10)\trans$), we also have reasonably good result.
 
\subsection{Application: background subtraction from surveillance video}
Video is a good candidate for low-rank subspace tracking due to the correlation between frames \citep{candes2011robust}. In surveillance video, background is generally stable and may change very slowly due to varying illumination. We experiment on both the airport and lobby surveillance video data which has been previously studied in \cite{li2004statistical,candes2011robust,he2012incremental}. To demonstrate that the OMWRPCA-CP algorithm can effectively track slowly changing subspace with change points, we consider panning a ``virtual camera'' moving from left to right and right to left through the video. The ``virtual camera'' moves at a speed of 1 pixel per 10 frames. The original frame has size $176\times 144$ and $160\times 128$ for the airport and the lobby video data respectively. The virtual camera has the same hight and half the width. We stack each frame to a column and feed it to the algorithms. To make the background subtraction task even more difficult, we add one change point to both video where the ``virtual camera'' jumps instantly from the most right-hand side to the most left-hand side. 

\begin{figure}[ht]
	\centering
	\subfloat[Original frames]{
		\includegraphics[width=0.16\textwidth]{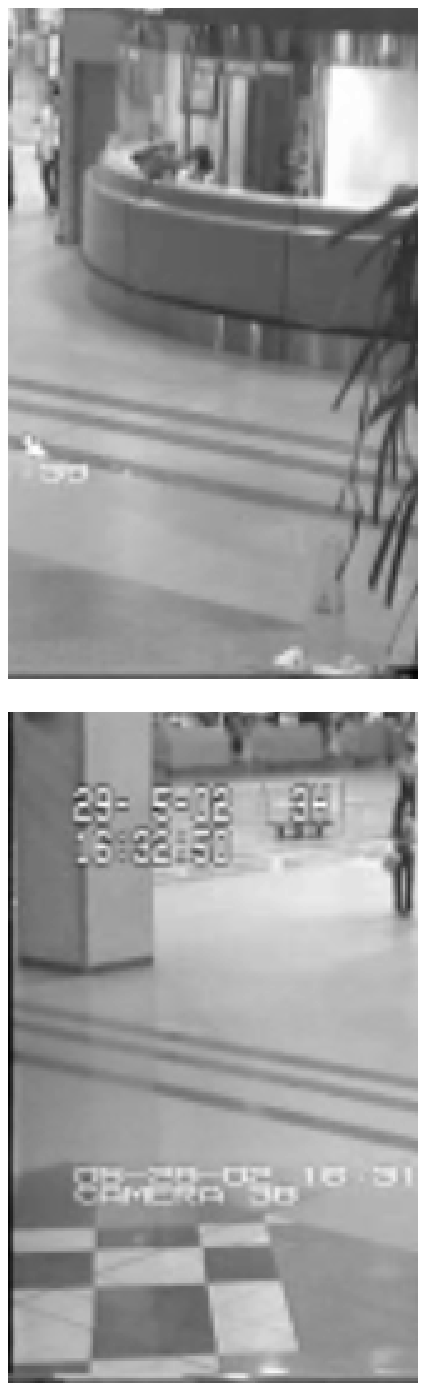}
		\label{fig:subfig1}}
	\quad
	\subfloat[Low-rank $\hat{\Lbf}$]{
		\includegraphics[width=0.16\textwidth]{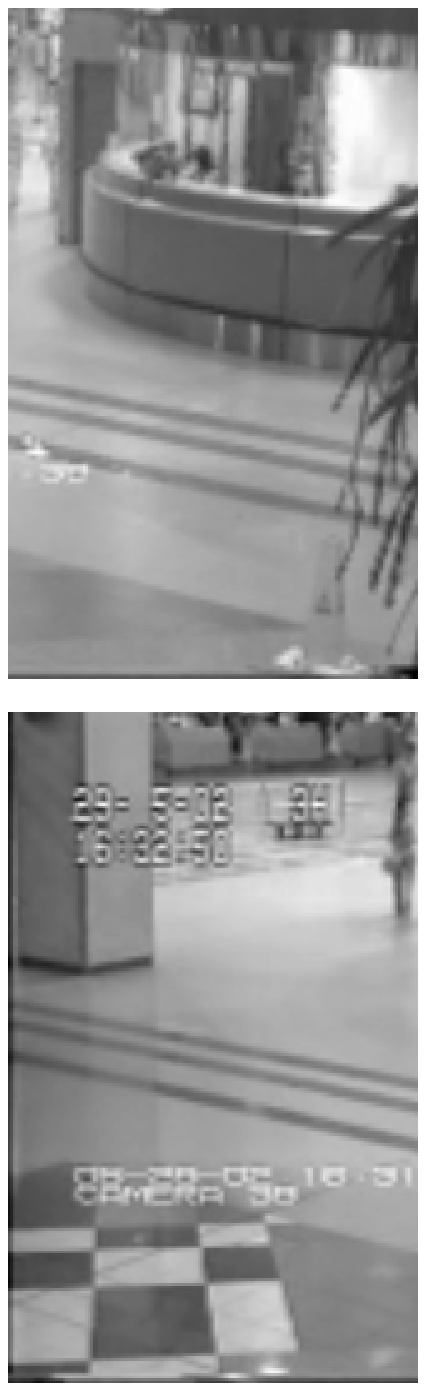}
		\label{fig:subfig2}}
	\quad
	\subfloat[Sparse $\hat{\Sbf}$]{
		\includegraphics[width=0.16\textwidth]{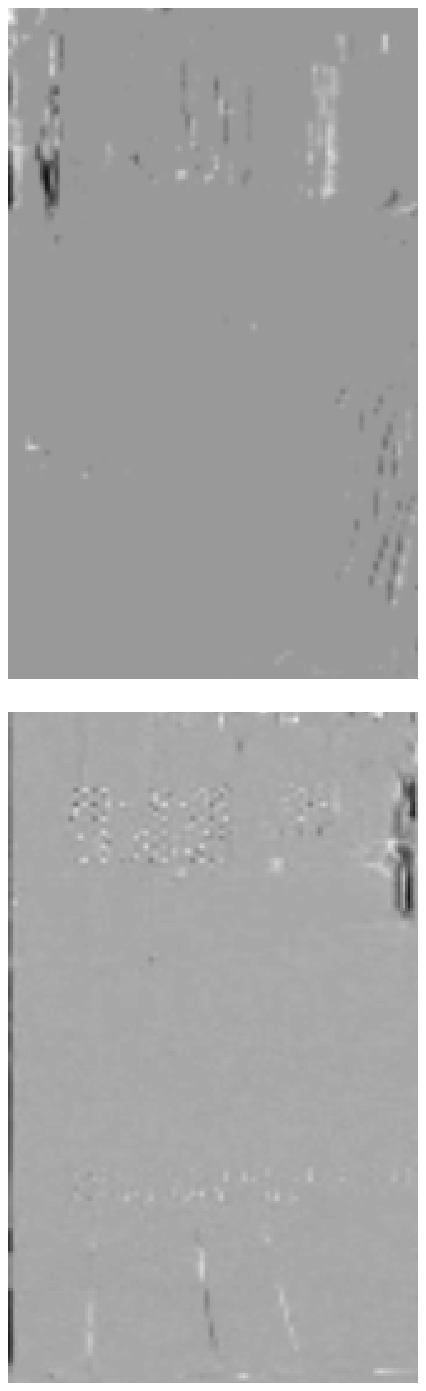}
		\label{fig:subfig3}}
	\quad
	\subfloat[Low-rank $\hat{\Lbf}$]{
		\includegraphics[width=0.16\textwidth]{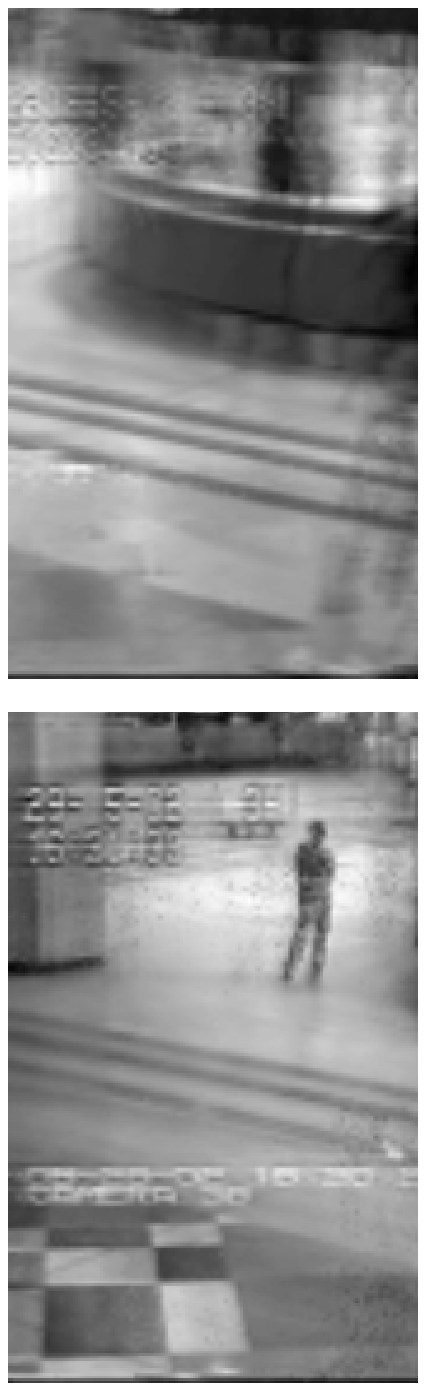}
		\label{fig:subfig4}}
	\quad
	\subfloat[Sparse $\hat{\Sbf}$]{
		\includegraphics[width=0.16\textwidth]{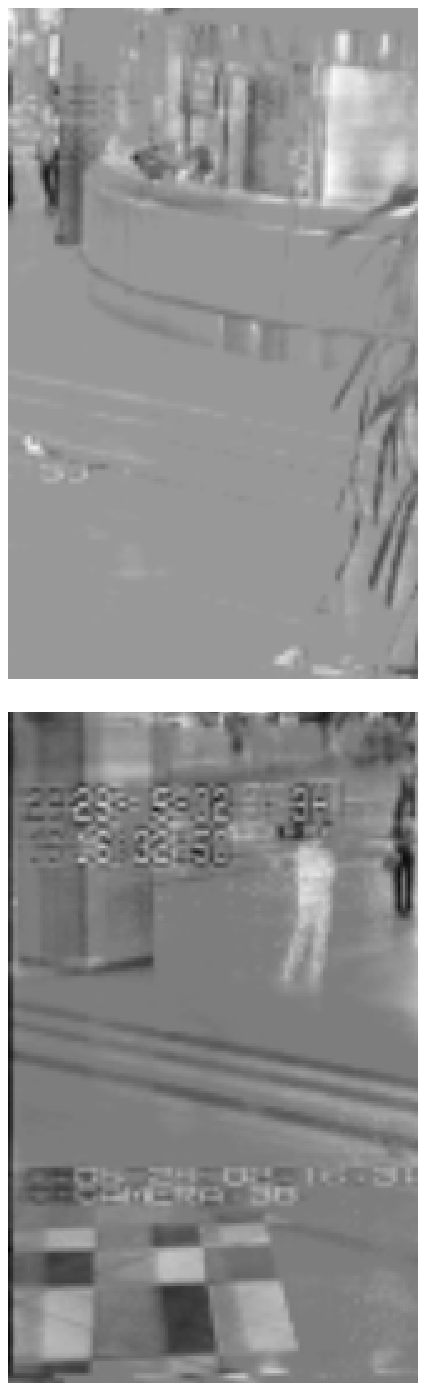}
		\label{fig:subfig5}}
	\caption{Background modeling from airport video. The first row shows the result at $t=878$. The second row shows the result at $t=882$. The change point is at $t=880$. (a) Original video $\Mbf$. (b)-(c) Low-rank $\hat{\Lbf}$ and Sparse $\hat{\Sbf}$ obtained from OMWRPCA-CP. (d)-(e)  Low-rank $\hat{\Lbf}$ and Sparse $\hat{\Sbf}$ obtained from STOC-RPCA.}
	\label{fig:airport}
\end{figure}

\begin{figure}[ht]
	\centering
	\subfloat[Original frames]{
		\includegraphics[width=0.16\textwidth]{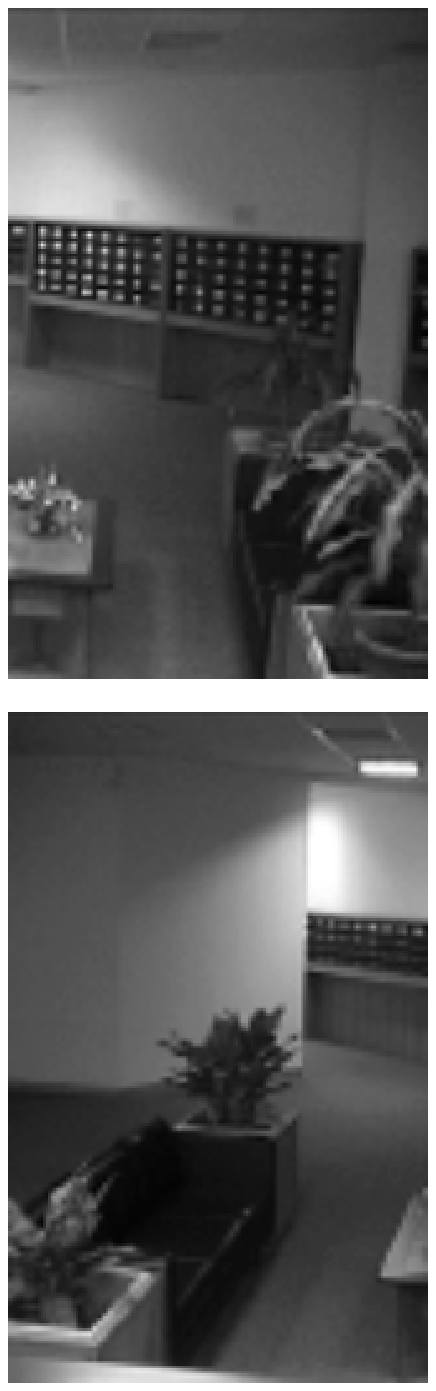}
		}
	\quad
	\subfloat[Low-rank $\hat{\Lbf}$]{
		\includegraphics[width=0.16\textwidth]{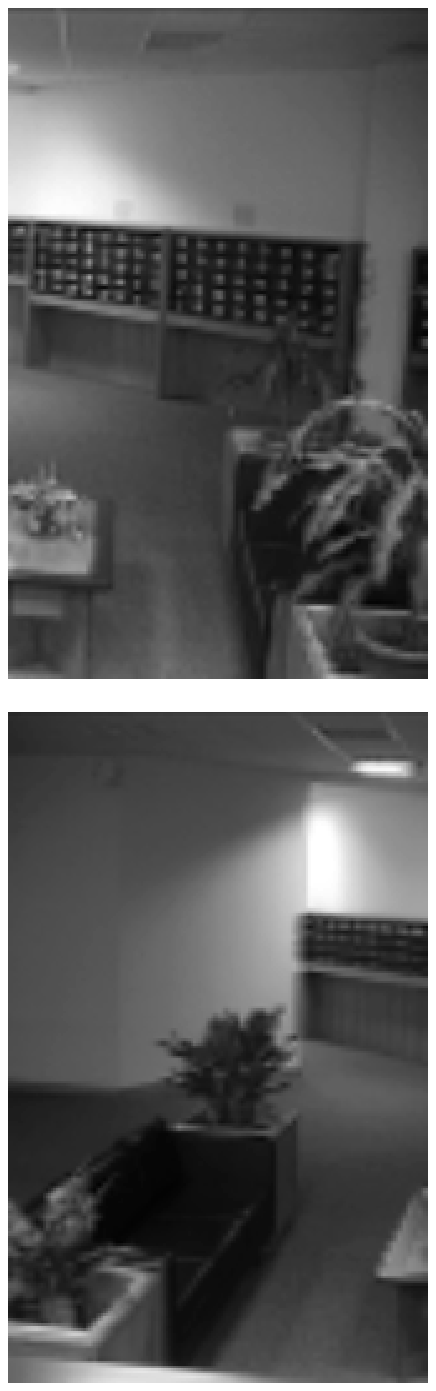}
		}
	\quad
	\subfloat[Sparse $\hat{\Sbf}$]{
		\includegraphics[width=0.16\textwidth]{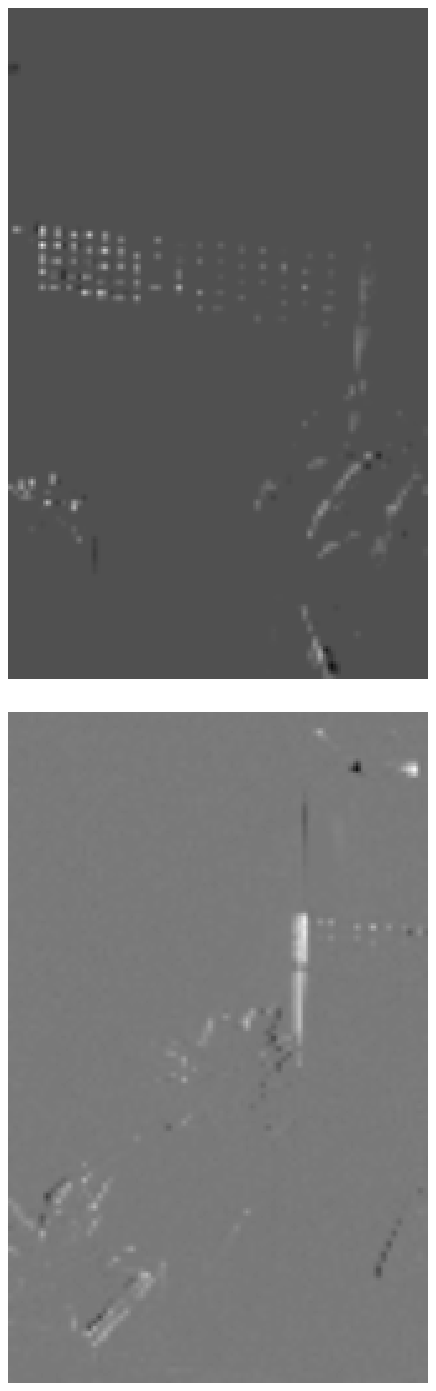}
		}
	\quad
	\subfloat[Low-rank $\hat{\Lbf}$]{
		\includegraphics[width=0.16\textwidth]{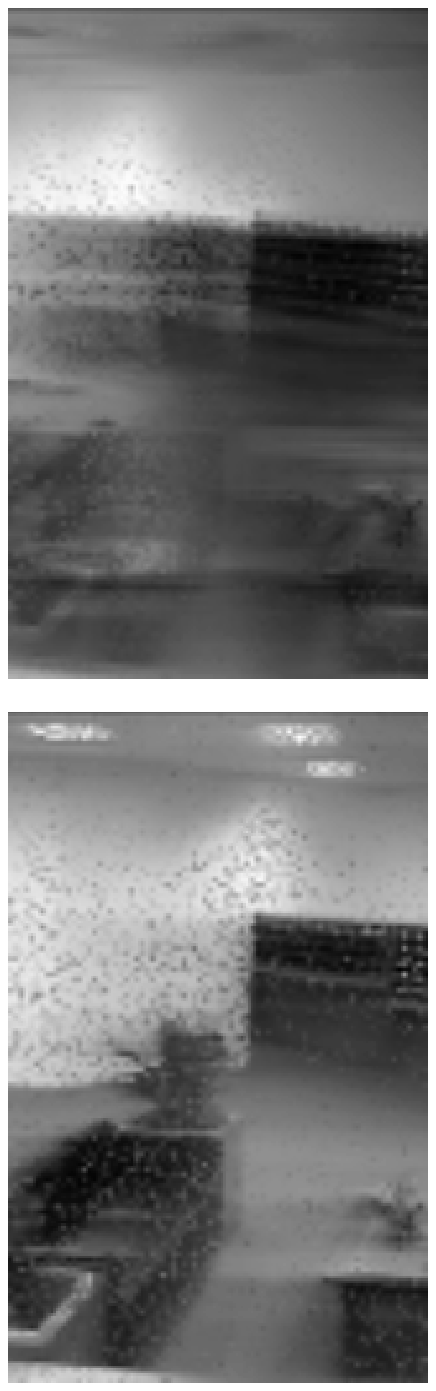}
		}
	\quad
	\subfloat[Sparse $\hat{\Sbf}$]{
		\includegraphics[width=0.16\textwidth]{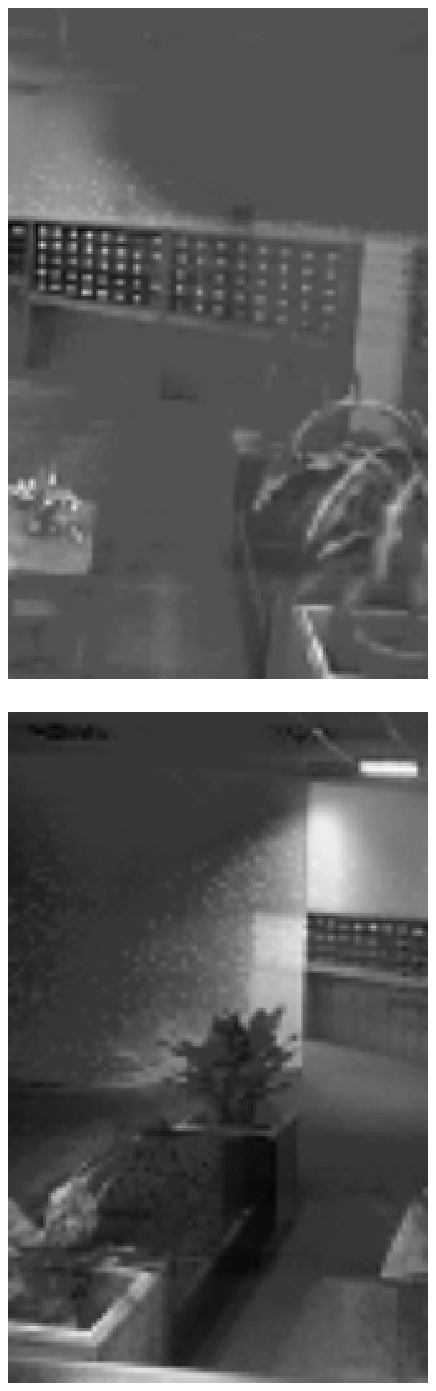}
		}
	\caption{Background modeling from lobby video. The first row shows the result at $t=798$. The second row shows the result at $t=802$. The change point is at $t=800$. (a) Original video $\Mbf$. (b)-(c) Low-rank $\hat{\Lbf}$ and Sparse $\hat{\Sbf}$ obtained from OMWRPCA-CP. (d)-(e)  Low-rank $\hat{\Lbf}$ and Sparse $\hat{\Sbf}$ obtained from STOC-RPCA.}
	\label{fig:lobby}
\end{figure}

We choose the following parameters in the algorithm, $\lambda_1=1/\sqrt{200}$, $\lambda_2=100/\sqrt{200}$, $n_{\mathrm{win}}=20$, $n_{\mathrm{burnin}}=100$, $n_{\mathrm{cp-burnin}}=100$, $n_{\mathrm{test}}=300$, $N_{\mathrm{check}}=3$, $\alpha_{\mathrm{prop}}=1$, $\alpha=0.01$, $n_{\mathrm{positive}}=3$ and $n_{\mathrm{tol}}=1000$. OMWRPCA-CP catches the true change points exactly in both experiments. We show the recovered low-rank $\hat{\Lbf}$ and sparse $\hat{\Sbf}$ at two frames before and after the change points in Figure~\ref{fig:airport} and Figure~\ref{fig:lobby} for airport and lobby video data, respectively. OMWRPCA-CP has much sharper recovered low-rank $\hat{\Lbf}$ compared with STOC\_RPCA. OMWRPCA-CP also has better performance in recovering sparse $\hat{\Sbf}$.

\section{Conclusion and Discussion}

In this paper we have proposed an online robust PCA algorithm. The algorithm can track both slowly and abruptly changed subspaces. By embedding hypothesis tests in the algorithm, the algorithm can discover the exact locations of change points for the underlying low-rank subspaces. Though in this work we have only applied the algorithm for real-time video layering where we separate the video sequence into a slowly changing background and a sparse foreground, we believe the algorithm can also be used for applications such as failure detection in mechanical systems, intrusion detection in computer networks and human activity recognition based on sensor data. These are left for future work. Another important direction is to develop an automatic (data-driven) method to choose tuning parameters in the algorithm.

\bibliography{omwrpca}
\bibliographystyle{ieeetr}

\end{document}